\documentclass[10pt,twocolumn,letterpaper]{article}

\usepackage{cvpr}              

\usepackage{graphicx}
\usepackage{amsmath}
\usepackage{amssymb}
\usepackage{booktabs}

\usepackage{float}
\usepackage{multirow}

\usepackage{animate}

\usepackage[ruled,linesnumbered]{algorithm2e}

\usepackage[accsupp]{axessibility}

\usepackage[pagebackref,breaklinks,colorlinks]{hyperref}

\usepackage[capitalize]{cleveref}
\usepackage{color}
\usepackage{makecell}
\newcommand{\CQ}[1]{{\textcolor{black}{ #1}}}

\crefname{section}{Sec.}{Secs.}
\Crefname{section}{Section}{Sections}
\Crefname{table}{Table}{Tables}
\crefname{table}{Tab.}{Tabs.}

\def\cvprPaperID{4247} 
\def\confName{CVPR}
\def\confYear{2023}

\begin{document}

\title{{CAP-VSTNet}: Content Affinity Preserved Versatile Style Transfer}

\author{
Linfeng Wen\textsuperscript{1}\quad
Chengying Gao\textsuperscript{1~*}\quad
Changqing Zou\textsuperscript{2,3}\\
\textsuperscript{1}Sun Yat-Sen University\quad \textsuperscript{2} State Key Lab of CAD\&CG, Zhejiang University \quad
\textsuperscript{3} Zhejiang Lab\\
{\tt\small wenlf5@mail2.sysu.edu.cn\quad mcsgcy@mail.sysu.edu.cn\quad aaronzou1125@gmail.com}
}

\twocolumn[{%
\renewcommand\twocolumn[1][]{#1}%
\maketitle
\begin{center}
    \centering
    \captionsetup{type=figure}

    \includegraphics[width=1.0\linewidth, height=0.34\linewidth]{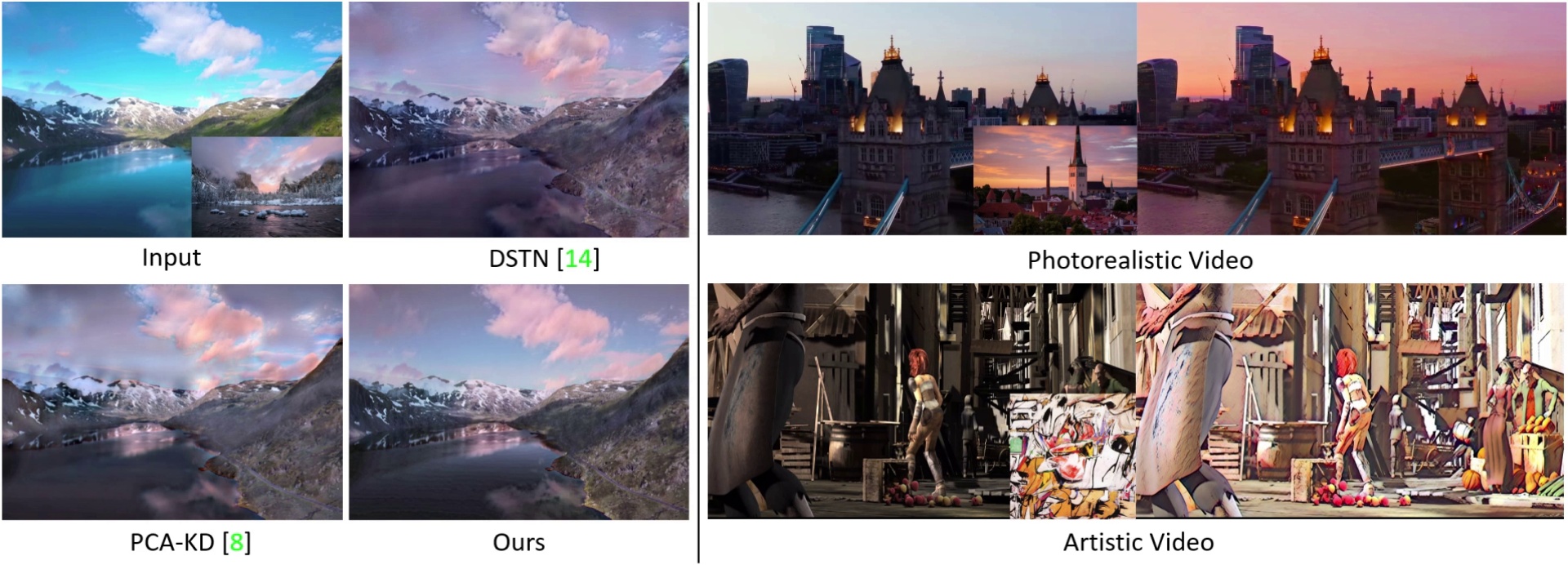}
    \captionof{figure}{Our model produces more consistent stylization results with content affinity preserved for photorealistic image style transfer (\textit{left}), and photorealistic video and artistic video style transfer  (\textit{right}). Animations can be found in the supplementary material.}
    \label{fig1-1}
\end{center}
}]

\begin{abstract}
Content affinity loss including feature and pixel affinity is a main problem which leads to artifacts in photorealistic and video style transfer. This paper proposes a new framework named CAP-VSTNet, which consists of a new reversible residual network and an unbiased linear transform module, for versatile style transfer. This reversible residual network can not only preserve content affinity but not introduce redundant information as traditional reversible networks, and hence facilitate better stylization. Empowered by Matting Laplacian training loss which can address the pixel affinity loss problem led by the linear transform, the proposed framework is applicable and effective on versatile style transfer. Extensive experiments show that CAP-VSTNet can produce better qualitative and quantitative results in comparison with the state-of-the-art methods. 
\footnotetext{*Corresponding Author}
\end{abstract}


\section{Introduction}
\label{sec:intro}

Photorealistic style transfer aims to reproduce content image with the style from a reference image in a photorealistic way. To be photorealism, the stylized image should preserve clear content detail and consistent stylization of the same semantic regions. 
Content affinity preservation, including feature and pixel affinity preservation~\cite{luan2017deep, li2018closed, li2019learning}, is the key to achieve both clear content detail and consistent stylization in the transfer.

The framework of a deep learning based photorealistic style transfer generally uses such an architecture: an encoder module extracting content and style features, followed by a transformation module to adjust features statistics, and finally a decoder module to invert stylized feature back to stylized image.
Existing photorealistic methods typically employ pre-trained VGG~\cite{simonyan2014very} as encoder. Since the encoder is specifically designed to capture object-level information for the classification task, it inevitably results in content affinity loss. To reduce the artifacts, existing methods either use skip connection modules~\cite{yoo2019photorealistic, an2020ultrafast, hong2021domain} or build a shallower network~\cite{li2019learning, wu2022ccpl, chiu2022pca}. However, these strategies, limited by the image recovery bias, cannot achieve a perfect content affinity preservation on unseen images.

In this work, rather than use the traditional encoder-transformation-decoder architecture, we resort to a reversible framework~\cite{an2021artflow} based solution called CAP-VSTNet, which consists of a specifically designed reversible residual network followed by an unbiased linear transform module based on Cholesky decomposition~\cite{kessy2018optimal} that performs style transfer in the feature space. 
The reversible network takes the advantages of the bijective transformation and can avoid content affinity information loss during forward and backward inference. However, directly using the reversible network cannot work well on our problem. This is because redundant information will accumulate greatly when the network channel increases. It will further lead to content affinity loss and noticeable artifacts as the transform module is sensitive to the redundant information. 
Inspired by knowledge distillation methods~\cite{wang2020collaborative, chiu2022pca}, we improve the reversible network and employ a channel refinement module to avoid the redundant information accumulation. We achieve this by spreading the channel information into a patch of the spatial dimension. In addition, we also introduce cycle consistency loss in CAP-VSTNet to make the reversible network robust to small perturbations caused by numerical error.

Although the unbiased linear transform based on Cholesky decomposition~\cite{kessy2018optimal} can preserve feature affinity, it cannot guarantee pixel affinity. Inspired by~\cite{luan2017deep, li2018closed}, we introduce Matting Laplacian~\cite{levin2007closed} loss to train the network and preserve pixel affinity. Matting Laplacian~\cite{levin2007closed} may result in blurry images
when it is used with another network like one with an encoder-decoder architecture. But it does not have this issue in CAP-VSTNet, since the bijective transformation of reversible network theoretically requires all information to be preserved.

CAP-VSTNet can be flexibly applied to versatile style transfer, including photorealistic and artistic image/video style transfer. We conduct extensive experiments to evaluate its performance. The results show it can produce better qualitative and quantitative results in comparison with the state-of-the-art image style transfer methods. We show that with minor loss function modifications, CAP-VSTNet can perform stable video style transfer and outperforms existing methods.

\section{Related Work}
\label{sec:related}

\subsection{Style Transfer}

Gatys et al.~\cite{gatys2016image} expose the powerful representation ability of deep neural networks and propose neural style transfer by matching the correlations of deep features. Feed-forward frameworks~\cite{johnson2016perceptual, ulyanov2016texture, wang2017multimodal} are proposed to address the issue of computational cost. To achieve universal style transfer, transformation modules are proposed to adjust statistics of deep features, such as the mean and variance~\cite{huang2017arbitrary} and the inter-channel correlation~\cite{li2017universal}. 

Photorealistic style transfer requires that stylized image should be undistorted and consistently stylized. DPST~\cite{luan2017deep} optimizes stylized image with regularization term computed on Matting Laplacian~\cite{levin2007closed} to suppress distortion. PhotoWCT~\cite{li2018closed} proposes post-processing algorithm by using Matting Laplacian as affinity matrix to reduce artifacts. However, both of these methods may blur the stylized images instead of preserving the pixel affinity. The following works~\cite{yoo2019photorealistic, an2020ultrafast, chiu2022pca} mainly focus on preserving clear details and speeding up processing by designing skip connection module or shallower network. Content affinity preservation including feature and pixel affinity preservation remains an unsolved challenge.

Recently, versatile style transfer has received a lot of attention. Many approaches focus on exploring a general framework capable of performing artistic, photorealistic and video style transfer.
Li et al.~\cite{li2019learning} propose a linear style transfer network and a spatial propagation network~\cite{liu2017learning} for artistic and photorealistic style transfer, respectively. DSTN~\cite{hong2021domain} introduces a unified architecture with domain-aware indicator to adaptively balance between artistic and photorealistic stylization. Chiu et al.~\cite{chiu2020iterative} propose an optimization-based method to achieve fast artistic or photorealistic style transfer by simply adjusting the number of iterations. Chen et al.~\cite{chen2021artistic} extend contrastive learning to artistic image and video style transfer by considering internal-external statistics. Wu et al.~\cite{wu2022ccpl} apply contrastive learning by incorporating neighbor-regulating scheme to preserve the coherence of the content source for artistic and photorealistic video style transfer. While achieving versatile style transfer, VGG-based networks suffer from inconsistent stylization due to content affinity loss. We show that preserving content affinity can improve image consistent stylization and video temporal consistency.

\begin{figure*}[ht]
  \centering
  \includegraphics[width=1.0\linewidth]{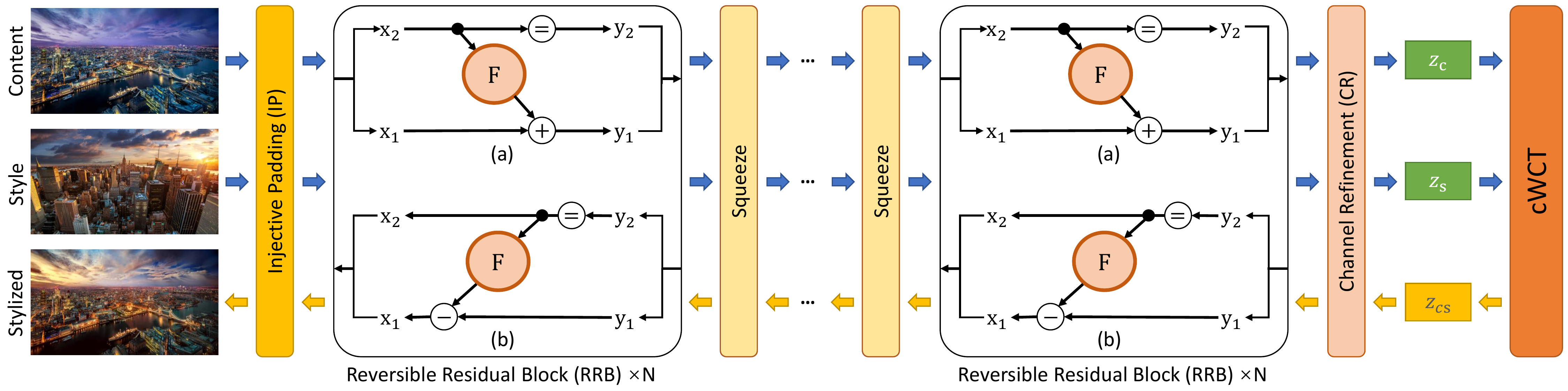}
  \caption{Architecture illustration of the proposed CAP-VSTNet. See Section~3 for details.}
  \label{fig3-1}
\end{figure*}

\begin{figure}
  \centering
  \includegraphics[width=0.85\linewidth]{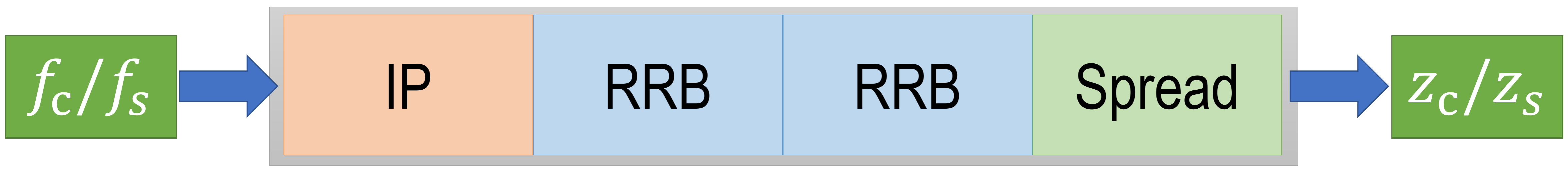}
  \caption{Structure of the adopted
  reversible residual blocks based channel refinement module (called CR-RRB for short). IP and RRB denotes injective padding module and reversible residual block, respectively.}
  \label{fig3-3}
\end{figure}

\subsection{Reversible Network}
Dinh et al.~\cite{dinh2014nice} first propose an estimator that learns a bijective transform between data and latent space, which can be seen as a perfect auto-encoder pair as it naturally satisfies reconstruction term of auto-encoder~\cite{bengio2007scaling, vincent2010stacked}. Follow-up work by Dinh et al.~\cite{dinh2016density} introduces new transformation that breaks the unit determinant of Jacobian to address volume-preserving mapping. 
Glow~\cite{kingma2018glow} proposes a simple type of generative flow building on the works by Dinh~\cite{dinh2014nice,dinh2016density}. 
Since each layer's activation of reversible network can be exactly reconstructed from the next layer's, RevNet~\cite{gomez2017reversible} and Reformer~\cite{kitaev2020reformer} present reversible residual layers to address memory consumption during deep network training. i-RevNet~\cite{jacobsen2018revnet} builds an invertible type of RevNet with invertible down-sampling module. i-ResNet~\cite{behrmann2019invertible} inverts residual mapping by using Banach fixed point theorem to address the restriction of reversible network architecture. 

Recently, An et al.~\cite{an2021artflow} apply flow-based model~\cite{kingma2018glow} to address the content leak problem for artistic style transfer. However, content affinity may not be preserved due to transformation module and redundant information, which leads to noticeable artifacts. The proposed method addresses this issue via a new reversible residual network enhanced by a channel refinement module and a Matting Laplacian loss based training.


\section{Method}
\label{method}
The architecture of CAP-VSTNet is shown in Figure~\ref{fig3-1}. Given a content image and a style image, our framework first maps the input content/style images to latent space through the forward inference of network after
an injective padding module which increases the input dimension by zero-padding along the channel dimension. 
The forward inference is performed
through cascaded reversible residual blocks and spatial squeeze modules. 
After that a channel refinement module is then used to remove the channel redundant information in content/style
image features for a more effective style transformation. Then a linear transform module cWCT is used to transfer the content representation to match the statistics of the style representation. Lastly the stylized representation is inversely mapped back to the stylized image through the backward inference. 

\subsection{Reversible Residual Network}

In our network design, each reversible residual block performs a function of a pair of inputs $x_1,x_2 \rightarrow y_1,y_2$, which can be expressed in the form:
\begin{equation}
\begin{split}
  x_1, x_2~&~=~~split(x), \\
  y_1=x_1 + &F(x_2), ~~~y_2=x_2.
\end{split}
\end{equation}
Following Gome et al. \cite{gomez2017reversible}, we use the channel-wise partitioning scheme that divides the input into two equal-sized parts along the channel dimension. Since the reversible residual block processes only half of the channel dimension at one time, it is necessary to perturb the channel dimension of the feature maps. We find that channel shuffling is effective and efficient: $ y = (y_2, y_1) $. Each block can be reversed by subtracting the residuals:
\begin{equation}
\begin{split}
  y_2, y_1~&~=~~split(y), \\
  x_2=y_2,  ~~~&x_1=y_1 - F(x_2).
\end{split}
\end{equation}
Figure~\ref{fig3-1} (a) and (b) show the illustration of the forward and backward inference of reversible residual block, respectively. The residual function $F$ is implemented by consecutive conv layers with kernel size 3. And each conv layer is followed by a relu layer, except for the last. We attain large receptive field by stacking multiple layers and blocks, in order to capture dense pairwise relations. We abandon the normalization layer as it poses a challenge to learn style representation.
To capture large scale style information, the squeeze module is used to reduce the spatial information by a factor of 2 and increase the channel dimension by a factor of 4. We combine reversible residual blocks and squeeze modules to implement a multi-scale architecture.

\begin{figure*}
  \centering
  \includegraphics[width=1.0\linewidth]{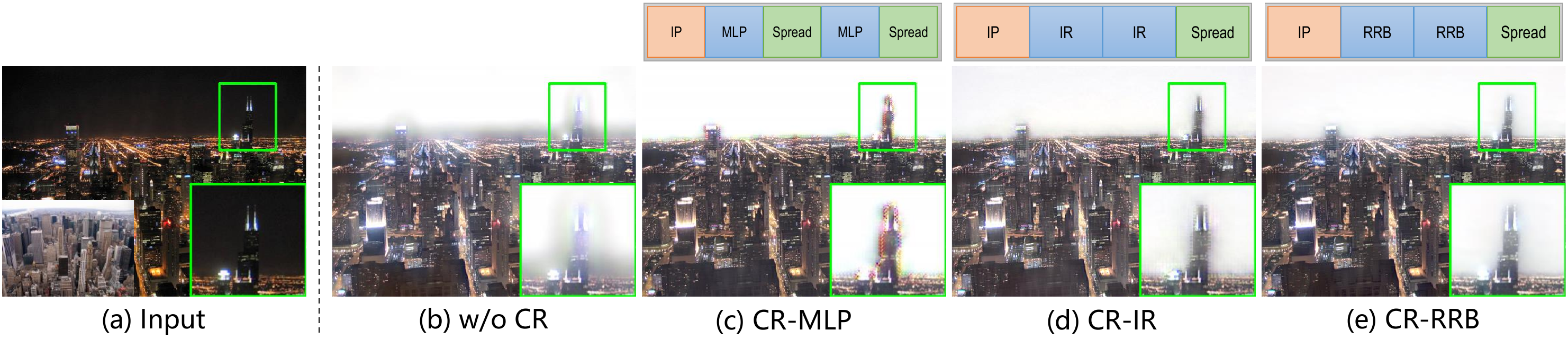}
  \caption{Alternative designs of the channel refinement module. (b) Artifacts are produced without any explicit redundant information removing. (c-d) Pointwise layers like fully connected layers or inverted residual layers cannot preserve good content affinity and result in aliasing artifacts. Please zoom in to see the details. The structures of the alternative designs are shown on top of the resulting images.}
  \label{fig3-2}
\end{figure*}

\subsection{Channel Refinement}
 The cascaded reversible residual block and squeeze module design in CAP-VSTNet
leads to redundant information accumulation during forward inference as the squeeze module exponentially increases the channels. 
The redundant information will negatively affect the stylization.  
In~\cite{wang2020collaborative, chiu2022pca}, 
channel compression is used to address
the redundant information problem and 
facilitate better stylization.
In our network design, we instead use a channel refinement module (CR) which is 
more suitable for the connected \CQ{cascaded reversible residual blocks}. 

As illustrated in Figure~\ref{fig3-3}, the CR module first uses an injective padding module increasing latent dimension to ensure that the input content/style image feature channel can be divisible by the target channel. Then, it uses patch-wise reversible residual blocks to integrate large-field information, after that it spreads the channel information into a patch of the spatial dimension. 
There are several alternative design choices for this 
CR module. MLP based pointwise layers can also be used for information distillation. Our preliminary experiments have found aliasing artifacts may appear when pointwise layers (e.g.,  fully connected layers or inverted residuals~\cite{sandler2018mobilenetv2}) are employed (see the results produced by CR-MLP and CR-IR in Figure~\ref{fig3-2}), but the adopted CR-RRB design does not have this issue.

\begin{table}
    \centering
    \begin{tabular}{lccccc}
    \toprule
    \multicolumn{2}{c}{Method} & AdaIN & WCT & LinearWCT & cWCT \\
    \hline
    \multicolumn{2}{c}{Reversible} & \checkmark & \checkmark & & \checkmark \\
    \multicolumn{2}{c}{Stability}  & \checkmark & & \checkmark & \checkmark \\
    \multicolumn{2}{c}{Learning-free} &  & \checkmark & & \checkmark\\ \hline
    \multirow{2}{*}{Time} & C=32  & 0.066 & 1.186 & 0.288 & 0.097 \\
                          & C=256 & 0.424 & 3.205 & 2.419 & 0.808 \\
    \bottomrule
    \end{tabular}
    \caption{Design choices of the linear transformation module.
    The adopted cWCT is reversible, stable, and learning-free. The execution time is evaluated on C × 512 × 512 feature maps for 100 times.}
    \label{tab3-1}
\end{table}

\subsection{Transformation Module}
Existing photorealistic methods typically employ WCT~\cite{li2017universal} as transformation module, which contains whitening and coloring steps. Both of the above steps require the calculation of singular value decomposition (SVD). However, the gradient depends on the singular values $\sigma$ by calculating $\frac{1}{min_{\sigma(i \neq j)} \sigma_i^2 - \sigma_j^2 }$ .  If the covariance matrix of content (style) feature map $\Sigma_c=f_c f_c^T (\Sigma_s=f_s f_s^T)$ has the same singular values, or the distance between any two singular values is close to 0, the gradient becomes infinite. It will further cause the WCT module to fail and the model training to crash. 

We use an unbiased linear transform based on Cholesky decomposition~\cite{kessy2018optimal} to address this problem. 
The Cholesky decomposition is derivable with gradient depending on $\frac{1}{\sigma}$. It does not require that the two singular values are not equal as SVD, thus is more stable. To avoid overflow, we can regularize it with an identity matrix: $\hat\Sigma=\Sigma+\epsilon I$. Another advantage of Cholesky decomposition is that 
its computational cost is much lower than that of SVD. Therefore, the adopted Cholesky decomposition based
WCT (cWCT for short) is more stable and faster. We show the comparison of various linear transformation modules~\cite{huang2017arbitrary, li2017universal, li2019learning} in Table~\ref{tab3-1}.

\begin{figure}
  \centering
  \includegraphics[width=1.0\linewidth]{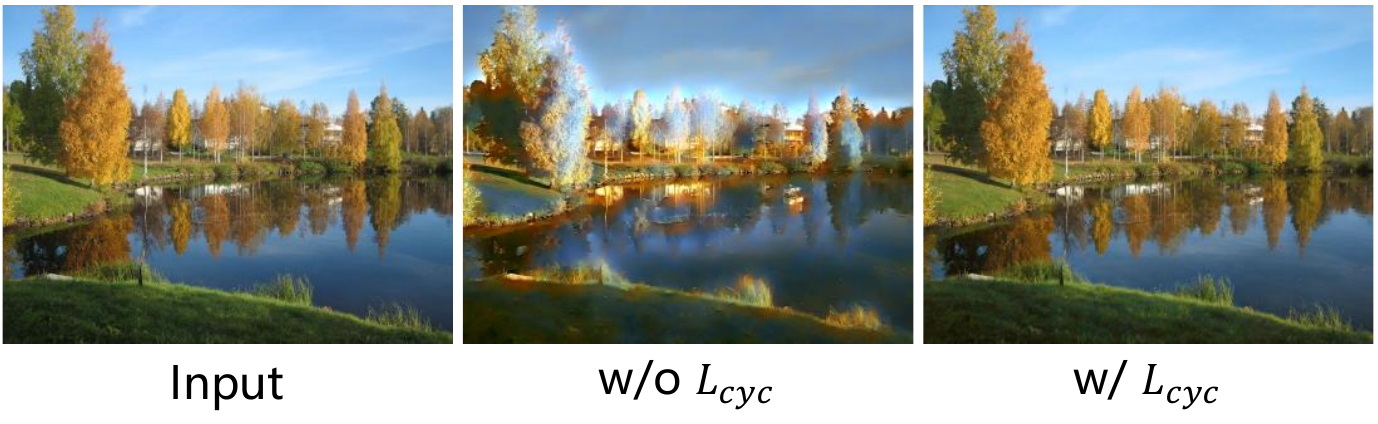}
  \caption{Ablation results of cycle consistency loss. Numerical error may results in significant changes.}
  \label{fig3-4}
\end{figure}

\begin{figure*}
  \centering
  \includegraphics[width=1.0\linewidth]{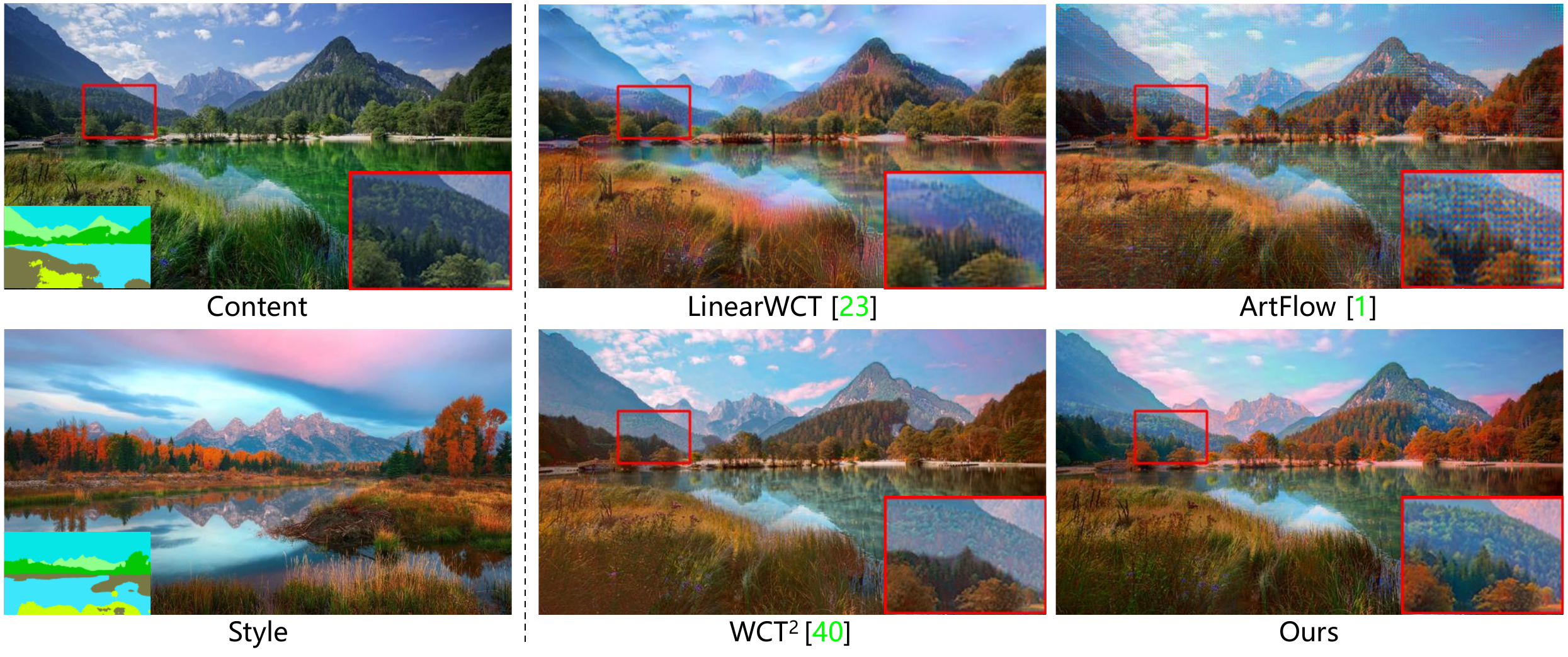}
  \caption{Visual comparison of content affinity preservation across various methods.}
  \label{fig4-1-3}
\end{figure*}

\begin{table*}
    \centering
    \begin{tabular}{cccccccc}
    \toprule
         & \textbf{Full} & w/o RRB & w/o IP & w/o CR-RRB & w/ CR-MLP & w/o $L_m\&L_{cyc}$ & w/o $L_{cyc}$ \\
        \midrule
        SSIM$\uparrow$ & \textbf{0.650} & 0.643 & 0.648 & 0.630 & 0.640 & 0.421 & 0.638 \\
        Gram loss$\downarrow$ & \textbf{0.750} & 0.811 & 0.874 & 0.831 & 0.803 & 0.873 & 0.782 \\
    \bottomrule
    \end{tabular}
    \caption{Quantitative comparison of different design choices in terms of structure preservation (SSIM) and stylization effect (Gram loss).}
    \label{tab4-1}
\end{table*}

\subsection{Training Loss}
 We train our network in an end-to-end manner with the integration of three types of losses:
\begin{equation}\label{EQtotalloss}
  L_{total} = L_s + \lambda_m L_m + \lambda_{cyc} L_{cyc},
\end{equation}
where $L_m$, $L_s$, and $L_{cyc}$
denote Matting Laplacian loss, style loss, and cycle consistency loss, respectively. $\lambda_m$ and $\lambda_{cyc}$ are the weights corresponding to the losses.

The Matting Laplacian loss in our design can be formulated as:  
\begin{equation}
  L_m = \frac{1}{N}\sum_{c=1}^{3} V_c[I_{cs}]^{T}MV_c[I_{cs}],
\end{equation}
where $N$ denotes the number of image pixels, $V_c [I_{cs}]$ denotes the vectorization of the stylized image $I_{cs}$ in channel c, and $M$ denotes the Matting Laplacian matrix of the content image $I_c$. 

Directly introducing Matting Laplacian loss in a network training could result in blurry images because Matting Laplacian loss will force the network to smooth the image rather than preserve pixel affinity.
Fortunately, introducing Matting Laplacian loss in our reversible network does not have the issue. It is because the bijective transformation in our reversible network requires all information to be preserved during forward and backward inference. 
The reversible network does not trick the loss by smoothing the image as it results in information loss. When performing linear transform, it depends on covariance matrix $\Sigma_s$. As the transformation of reversible network is deterministic, only a few style images with smooth texture may smooth the content structure. In this situation, it is reasonable to output a stylized image with the same smooth texture as we aim to transfer vivid style.

The style loss is formulated as:
\begin{equation}\label{EQstyleloss}
\begin{split}
  L_s &= \sum_{i=1}^{l} ||\mu(\phi_i(I_{cs})) - \mu(\phi_i(I_s))|| \\
      &+ \sum_{i=1}^{l} ||\sigma(\phi_i(I_{cs})) - \sigma(\phi_i(I_s))||,
\end{split}
\end{equation}
where $I_s$ denotes style image, $\phi_i$ denotes the $i_{th}$ layer of the VGG-19 network (from $ReLu1\_1$ to $ReLu4\_1$), and $\mu$ and $\sigma$ denote the mean and variance of the feature maps, respectively. 

Since all modules are reversible, we should be able to cyclically reconstruct content image $\Tilde{I_C}$ by transferring the style information of content image $I_c$ to stylized image $I_{cs}$. 
However, the reversible network suffers from numerical error and may result in noticeable artifacts (Figure~\ref{fig3-4}). Thus, we introduce the cycle consistency loss to improve the network robustness.

The cycle consistency loss is calculated with $L1$ distance:
\begin{equation}
  L_{cyc} = ||\Tilde{I_C}-I_C||_{1} .
\end{equation}

\begin{figure*}
  \centering
  \includegraphics[width=1.0\linewidth]{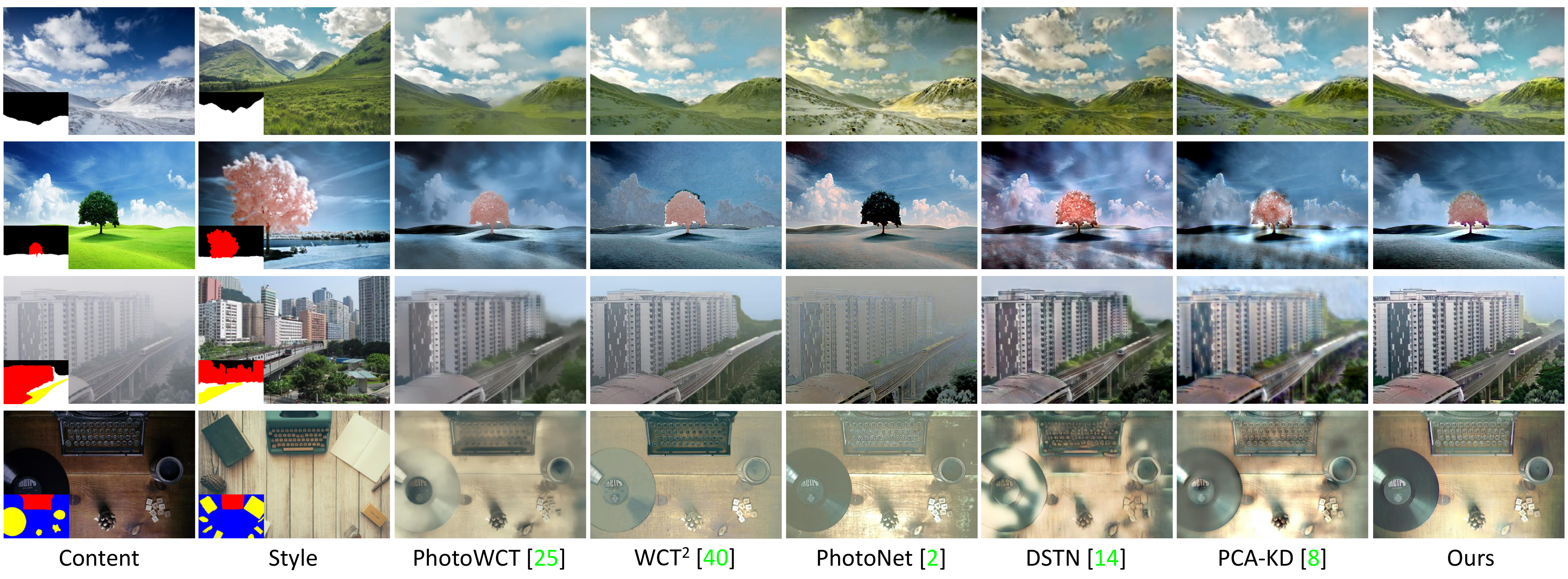}
  \caption{Visual comparisons of photorealistic image style transfer. All methods conduct style transfer with the assistance of masks, except PhotoNet which does not support masks.}
  \label{fig5-1}
\end{figure*}

\begin{table*}
  \centering
  \begin{tabular}{ccccccc}
    \toprule
    Method & PhotoWCT~\cite{li2018closed} & WCT$^2$~\cite{yoo2019photorealistic} & PhotoNet~\cite{an2020ultrafast} & DSTN~\cite{hong2021domain} & PCA-KD~\cite{chiu2022pca} & Ours \\
    \midrule
    SSIM$\uparrow$ & 0.582  &  \underline{0.644}  &  0.608  & 0.566  & 0.634 & \textbf{0.650}  \\
    Gram loss$\downarrow$  & 1.539  &  \underline{0.796}  &  1.970 & 0.996 & 1.162 & \textbf{0.750} \\
    Time$\downarrow$ & 16.88 & 0.32 & 0.19 & 0.92 & \textbf{0.05} & \underline{0.12} \\
    Parameters & 8.35M & 10.12M & 40.24M & 103.45M  & 334K & 4.09M \\
    \bottomrule
  \end{tabular}
  \caption{Quantitative comparison of photorealistic style transfer methods. The execution time is evaluated on 1024 × 512 resolution.}
  \label{table5-1}
\end{table*}

\subsection{Video Style Transfer}
Single-frame methods~\cite{li2019learning, yoo2019photorealistic, wu2022ccpl} show that applying image algorithms that operate on each video frame individually is possible. Since our framework preserves the affinity of input videos, which is naturally consistent and stable, the content of stylized video is also visually stable. To constrain the style of stylized video~\cite{gupta2017characterizing, li2019learning},
we have two strategies: adjust the style loss (Eq.\ref{EQstyleloss}) with lower layers of VGG-19 network (from $ReLu1\_1$ to $ReLu3\_1$) or 
add the regularization~\cite{wang2020regularization} to Eq.\ref{EQtotalloss} and fine-tune the model. Both strategies can achieve good temporal consistency. We choose the latter
one as it can produce slightly better stylization effect.


\section{Analysis}

\subsection{Content Affinity Preservation}
To show the advantages of preserving feature and pixel affinity, we compare the stylization results with three types of methods. As shown in Figure~\ref{fig4-1-3}, LinearWCT~\cite{li2019learning} applies linear transform to preserve feature affinity. However, the image details is unclear and the stylization is inconsistent as feature and pixel affinity could be damaged by VGG-base network. WCT$^2$~\cite{yoo2019photorealistic} aims to preserve spatial information rather than content affinity. While preserving clear details, it particularly relies on the precise masks, which otherwise produce noticeable seams. ArtFlow~\cite{an2021artflow} uses the flow-based model to address content leak problem. However, it typically generates noticeable artifacts as linear transform and redundant information damage content affinity. Compared with other methods, ours model not only preserves clear details, but also achieves seamless style transfer.

\subsection{Ablation Study}
We conduct an ablation study to quantitatively evaluate how much each component (i.e., channel refinement components and training losses) affects the visual effects. Table~\ref{tab4-1} shows the ablation study results. When all the design components are used, the network can obtain the best results. Replacing residual block (RRB) with inverted residuals~\cite{sandler2018mobilenetv2} degrades performance as the pointwise layer has smaller receptive field and damages content affinity. Removing injective padding (IP), the model fails to capture high-level content and style information from pixel image. Adding the channel refinement module (CR-RRB) helps remove redundant information for better content preservation and stylization effect. Implementing the channel refinement module with CR-MLP results in aliasing artifact, which degrades content affinity. 
Using VGG content loss (w/o $L_m\&L_{cyc}$) cannot guarantee pixel affinity due to the linear transform. 
With cycle consistency loss ($L_{cyc}$), the network achieves robustness to small perturbations. 

\begin{figure*}
  \centering
  \includegraphics[width=0.99\linewidth]{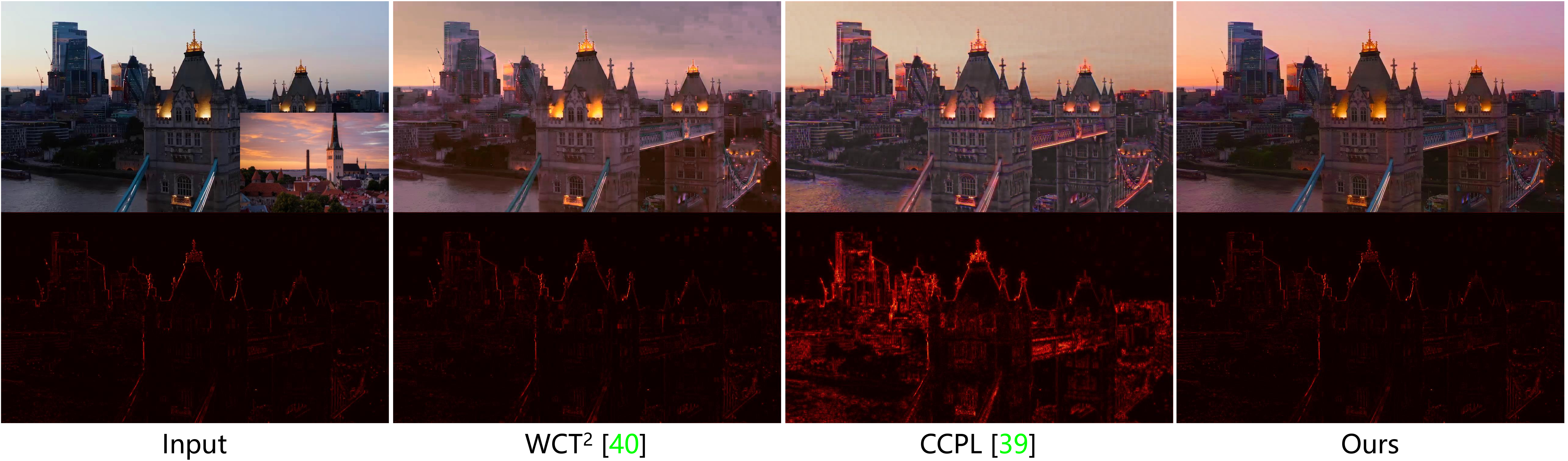}
  \caption{Comparisons of short-term temporal consistency on photorealistic video style transfer. The odd rows show the previous frame. The even rows show the temporal error heatmap.}
  \label{fig5-2}
\end{figure*}

\begin{figure*}
  \centering
  \includegraphics[width=0.99\linewidth]{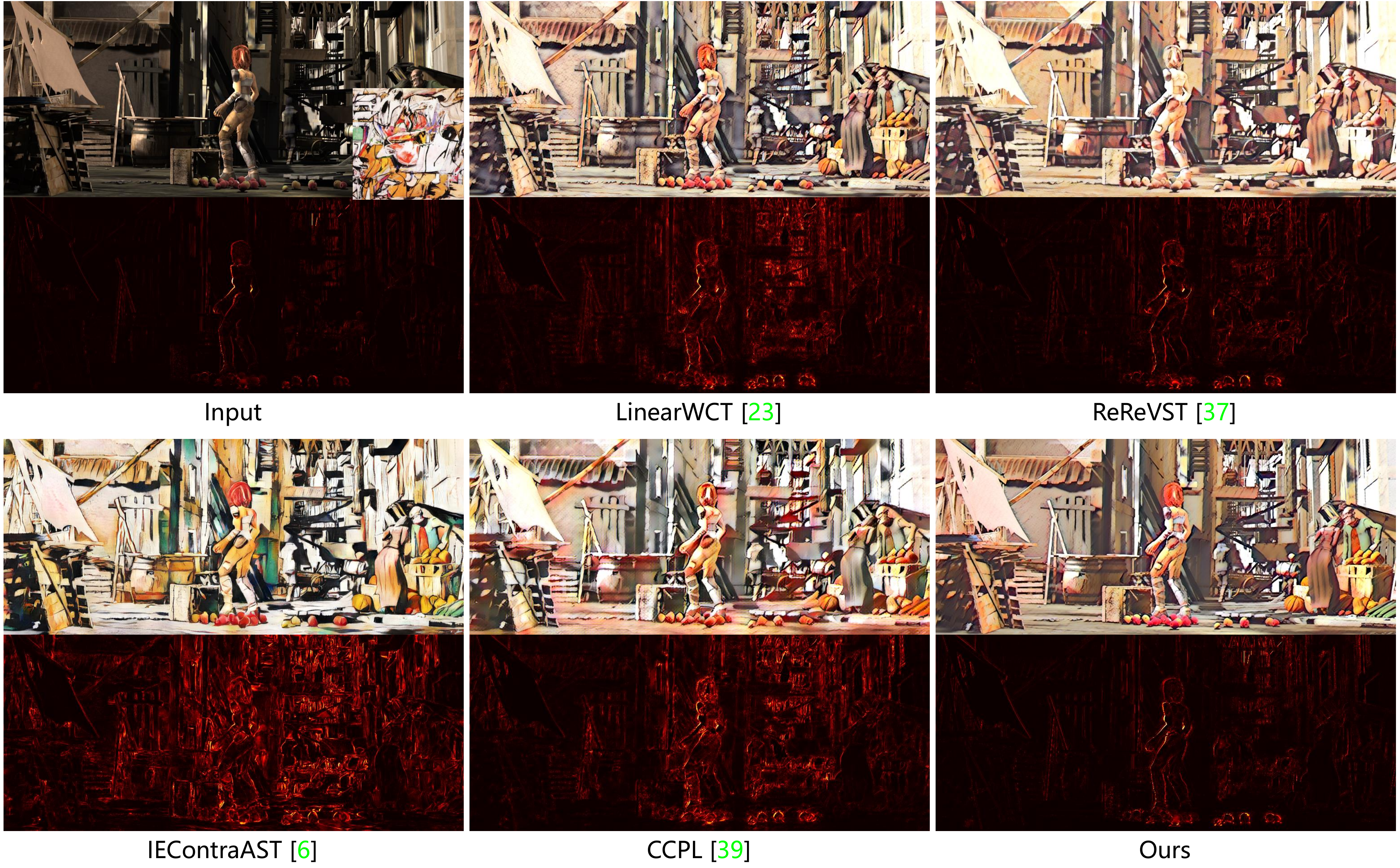}
  \caption{Comparisons of short-term temporal consistency on artistic video style transfer. The odd rows show the previous frame. The even rows show the temporal error heatmap.}
  \label{fig5-3}
\end{figure*}

\section{Experiments}

\subsection{Implementation Details}
We implement a three-scale architecture with 30 blocks and 2 squeeze modules. For photorealistic style transfer, we sample content and style images from MS COCO dataset~\cite{lin2014microsoft} and randomly crop them to 256$\times$256. 
We set the weight factors of loss function as: $\lambda_m=1200$ and $\lambda_{cyc}=10$. We train the network for 160,000 iterations using Adam optimizer with batch size of 2. The initial learning rate is set to 1e-4 and decays at 5e-5. For artistic style transfer, we set $\lambda_m=\lambda_{cyc}=1$ to allow more variation of image pixel and sample style images from WikiArt dataset~\cite{karayev2013recognizing}. All the experiments are conducted on a single NVIDIA RTX 3090 GPU.

\subsection{Photorealistic Image Style Transfer}
\paragraph{Qualitative evaluation.}
Figure~\ref{fig5-1} shows the comparison of the stylization results with advanced photorealistic style transfer methods, including PhotoWCT~\cite{li2018closed}, WCT$^2$~\cite{yoo2019photorealistic}, PhotoNet~\cite{an2020ultrafast}, DSTN~\cite{hong2021domain} and PCA-KD~\cite{chiu2022pca}. We can see that PhotoWCT usually generates blurry images with loss of details. Although WCT$^2$ faithfully preserves image spatial information, it produces noticeable seams. PhotoNet generates poor stylization effect due to discarding masks. DSTN stylizes images with noticeable artifacts and distorts image structure. PCA-KD is not able to produce consistent stylization. Compared with the existing methods, our method faithfully preserves image details and achieves better stylization effect. Besides, image stylization is consistent without artifacts, which greatly enhances photorealism.

\paragraph{Quantitative evaluation.}
Following previous works~\cite{yoo2019photorealistic, hong2021domain}, we use structural similarity (SSIM) to evaluate photorealism and Gram loss~\cite{gatys2016image} to evaluate stylization effect. We use all pairs of content and style images with semantic segmentation masks provided by DPST \cite{luan2017deep} for quantitative evaluation. Table~\ref{table5-1} shows the comparison of quantitative results. Our method not only preserves structure better, but also achieves stronger stylization effect. Since the reversible residual network naturally satisfies the reconstruction condition, we reduce network parameters and make it more lightweight than most of standard VGG-based networks. PCA-KD~\cite{chiu2022pca} applies knowledge distillation method to crate the lightweight model for ultra-resolution style transfer. We note that our model is also applicable for ultra-resolution (i.e., 4K resolution) and achieves better performance as well.

\begin{table}
\centering
\begin{tabular}{lcccc}
\hline
\multirow{2}{*}{Method} & \multirow{2}{*}{Gram loss$\downarrow$} & \multicolumn{2}{c}{Temporal loss$\downarrow$} \\ \cline{3-4}
                  &  & i=1  & i=10 \\ \hline
WCT$^2$~\cite{yoo2019photorealistic}  &  0.665 &  \underline{0.040} & \underline{0.108}  \\
CCPL~\cite{wu2022ccpl}  &  \underline{0.527} & 0.069 & 0.132 \\
Ours   & \textbf{0.435}  &  \textbf{0.039} &  \textbf{0.107} \\ \hline
\end{tabular}
\caption{Quantitative comparison of photorealistic video style transfer methods. 'i' denotes frame interval.}
\label{table5-3}
\end{table}

\begin{table}
\centering
\begin{tabular}{lcccc}
\hline
\multirow{2}{*}{Method} & \multirow{2}{*}{Gram loss$\downarrow$} & \multicolumn{2}{c}{Temporal loss$\downarrow$} \\ \cline{3-4}
                  &  & i=1  & i=10 \\ \hline
LinearWCT~\cite{li2019learning}  &  0.473 &  0.117 & 0.237 \\
ReReVST~\cite{wang2020consistent}  &  0.815 & \underline{0.108} & \underline{0.235}  \\
IEContraAST~\cite{chen2021artistic}  & 1.062 & 0.141 &  0.262   \\
CCPL~\cite{wu2022ccpl}  &  \textbf{0.371} & 0.128 & 0.251  \\
Ours   & \underline{0.436}  &  \textbf{0.104} & \textbf{0.228}  \\ \hline
\end{tabular}
\caption{Quantitative comparison of artistic video style transfer methods. 'i' denotes frame interval.}
\label{table5-2}
\end{table}

\subsection{Video Style Transfer}

\paragraph{Photorealistic video style transfer.}
We compare our method with state-of-the-art methods~\cite{yoo2019photorealistic, wu2022ccpl}. To visualize video stability, we show the heatmap of temporal error between the consecutive frames in Figure~\ref{fig5-2}. To quantitatively evaluate, we collect 20 pairs of video clips of multiple scenes and semantically related style images from the Internet. Following~\cite{wang2020consistent, wu2022ccpl}, we adopt the temporal loss to measure temporal consistency. We use RAFT~\cite{teed2020raft} to estimate the optical flow for short-term consistency (two adjacent frames) and long-term consistency (9 frames in between) evaluation. Table~\ref{table5-3} shows that our framework performs well against the other methods.

\paragraph{Artistic video style transfer.}
Figure~\ref{fig5-3} shows the comparison with four advanced methods~\cite{li2019learning, wang2020consistent, chen2021artistic, wu2022ccpl}. To quantitatively evaluate, we use all the sequences of MPI Sintel dataset~\cite{butler2012naturalistic} and collect 20 artworks of various types to stylize each video. For short-term consistency, MPI Sintel provides ground truth optical flows. For long-term consistency, we use PWC-Net~\cite{sun2018pwc} to estimate the optical flow following~\cite{wang2020consistent, wu2022ccpl}. Table~\ref{table5-2} shows that our framework achieves the best temporal consistency, thanks to the content affinity preservation. 
Our model also produces vivid stylization effect comparable to CCPL~\cite{wu2022ccpl}.

\subsection{Limitation}
\begin{figure}
  \centering
  \includegraphics[width=1.0\linewidth]{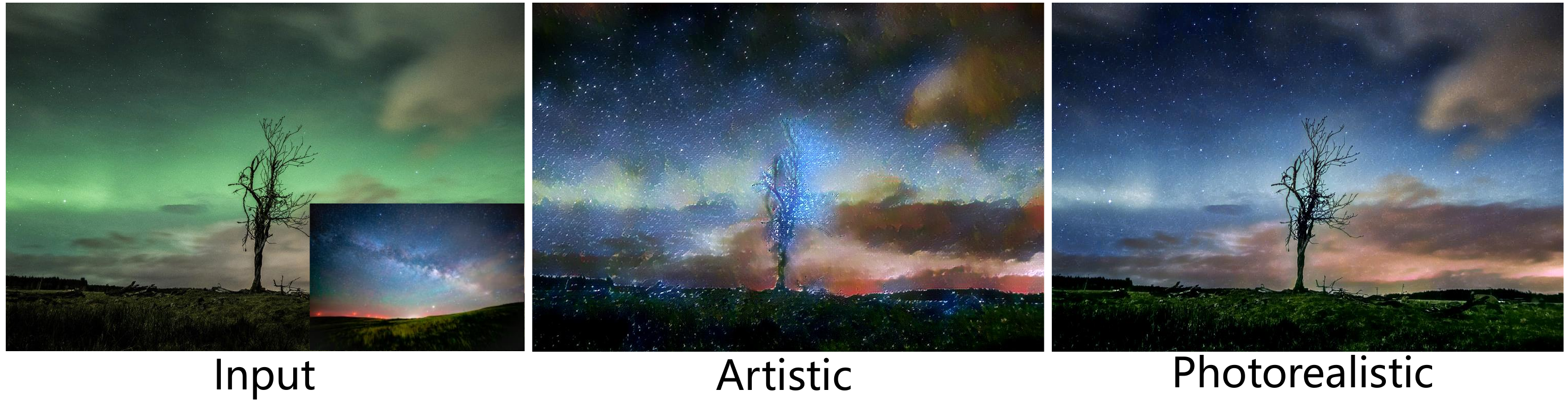}
  \caption{Limitation. Both our artistic and photorealistic models fail to transfer complex textures like milky way.}
  \label{fig5-4}
\end{figure}

Preserving content affinity helps to achieve consistent stylization. However, both our artistic and photorealistic models fail to capture complex texture and may generate artifacts (Figure~\ref{fig5-4}). Generating realistic textures remains a challenge for style transfer and image generation tasks. Existing stylization methods typically build on small models (e.g., VGG). Since realistic texture requires much high-frequency details, an interesting direction is to investigate whether large models can solve this problem.

\section{Conclusion}
In this paper, we propose a new framework named CAP-VSTNet for versatile style transfer, which consists of a new effective reversible residual network and an unbiased linear transform. It can preserve
two major content affinity: pixel and feature affinity
with the introduction of Matting Laplacian training loss.
We show that CAP-VSTNet achieves consistent and vivid stylization with clear details. CAP-VSTNet is also flexible for photorealistic and artistic video style transfer. Extensive experiments demonstrate the effectiveness and superiority of CAP-VSTNet in comparisons with state-of-the-art approaches.


\section*{Acknowledgement}
This work was supported by the Natural Science Foundation of Guangdong Province, China (Grant No. 2022A1515011425).



\clearpage

\appendix

\renewcommand\thesection{\Alph{section}}

\SetKw{Data}{Require}
\SetKwInput{kwInit}{Optional}


\section{Pseudocode of CAP-VSTNet}
\begin{algorithm}
\caption{Video style transfer process of CAP-VSTNet}
\KwIn{content frames ${\{I_c^j\}}_{j=1}^N$, style image $I_s$, the proposed new framework CAP-VSTNet which consists an unbiased style transfer module cWCT;}
\kwInit{content semantic masks ${\{M_c^j\}}_{j=1}^N$, style semantic mask $M_s$;}
\KwResult{stylized frames ${\{I_{cs}^j\}}_{j=1}^N$;}

feed $I_s$ to CAP-VSTNet and perform the forward inference of CAP-VSTNet to obtain the style feature $f_s$\;
\For{$j\leftarrow 1$ \KwTo $N$}{
  feed $I_c^j$ to CAP-VSTNet and perform the forward inference of CAP-VSTNet to obtain the content feature $f_c^j$\;
  \lIf{semantic masks $M_c^j$ and $M_s$ are provided}{
    feed $f_c^j, f_s, M_c^j, M_s$\ to cWCT and obtain the stylized feature $f_{cs}^j$\
  }
  \lElse{feed $f_c^j, f_s$\ to cWCT and obtain the stylized feature $f_{cs}^j$}\
  perform the backward inference of CAP-VSTNet to obtain the stylized frame $I_{cs}^j$\;
}
\end{algorithm}


\section{Unbiased Transformation Module}
We show that the whitening and coloring transforms in
cWCT is an unbiased style transfer module. Suppose we have a style transfer module $f_{cs} = C(f_c)S(f_s)$, where $C, S$ denote the content and style factor, and $f_c, f_s$ denote the content and style feature. ArtFlow defines that $f_{cs}$ is an unbiased style transfer moudle if $C(f_{cs})=C(f_c)$ and $S(f_{cs})=S(f_s)$.

Without loss of generality, we assume both $f_c$ and $f_s$ are centered. We have,
\begin{equation}
\begin{split}
  Whitening&: ~\hat{f_c} = L_c^{-1}f_c, \\
  Coloring&:  f_{cs} = L_s\hat{f_c}.
\end{split}
\end{equation}
where $f_c f_c^T=L_c L_c^T$, $f_s f_s^T=L_s L_s^T$ and $L$ is a triangular matrix. Therefore,
\begin{equation}
  f_{cs} = L_s L_c^{-1}f_c.
\end{equation}
Here,
\begin{equation}
  C(f) = L^{-1}f,  S(f) = L.
\end{equation}
Since,
\begin{equation}
  f_{cs} f_{cs}^T = f_s f_s^T = L_s L_s^T,
\end{equation}
We have,
\begin{equation}
  C(f_{cs}) = L_s^{-1}f_{cs} = L_c^{-1}f_c = C(f_c),
\end{equation}
\begin{equation}
  S(f_{cs}) = L_s = S(f_s).
\end{equation}
Therefore, cWCT is unbiased.


\section{Style Interpolation}
\begin{figure*}
  \centering
  \includegraphics[width=1.0\linewidth]{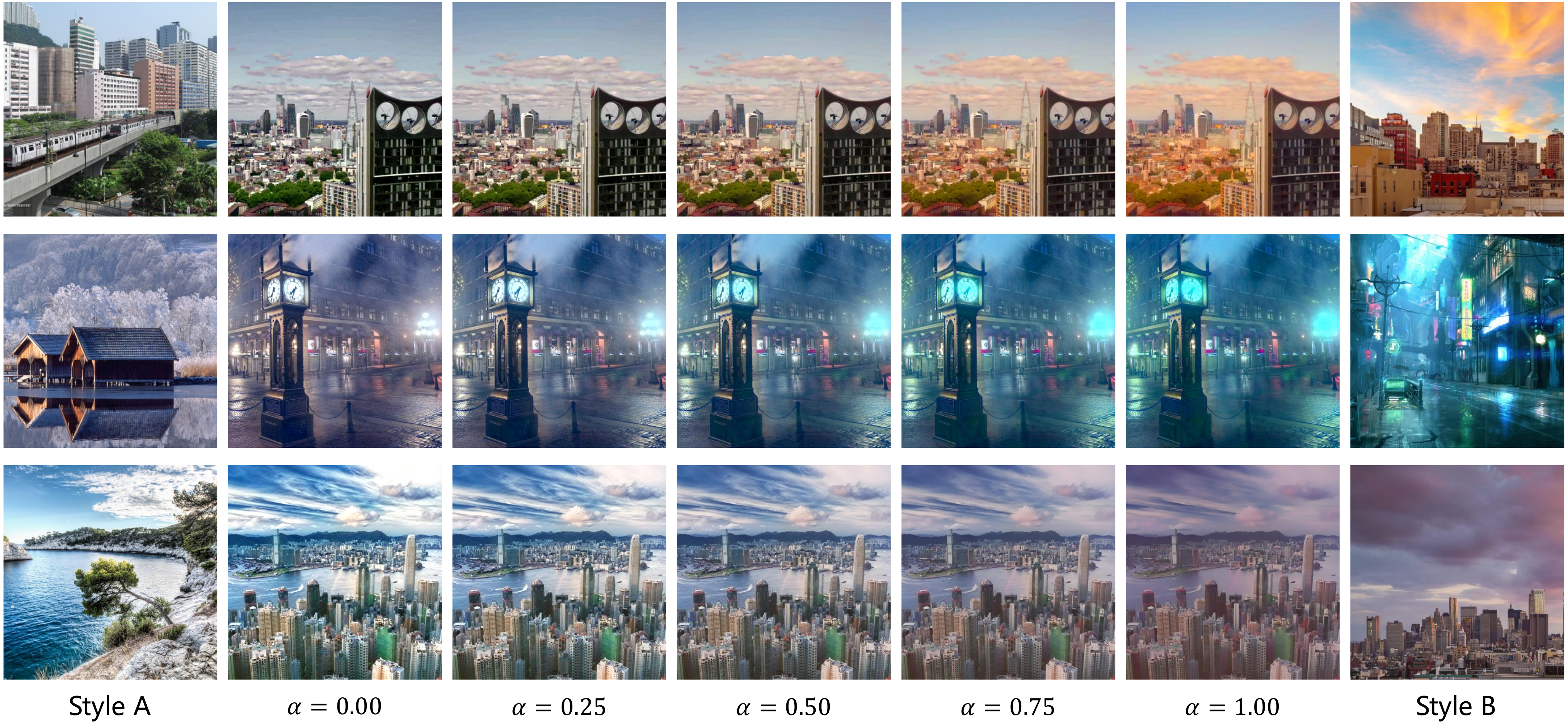}
  \caption{Style interpolation results of CAP-VSTNet on photorealistic style transfer.}
  \label{sp3-1-1}
\end{figure*}

\begin{figure*}
  \centering
  \includegraphics[width=1.0\linewidth]{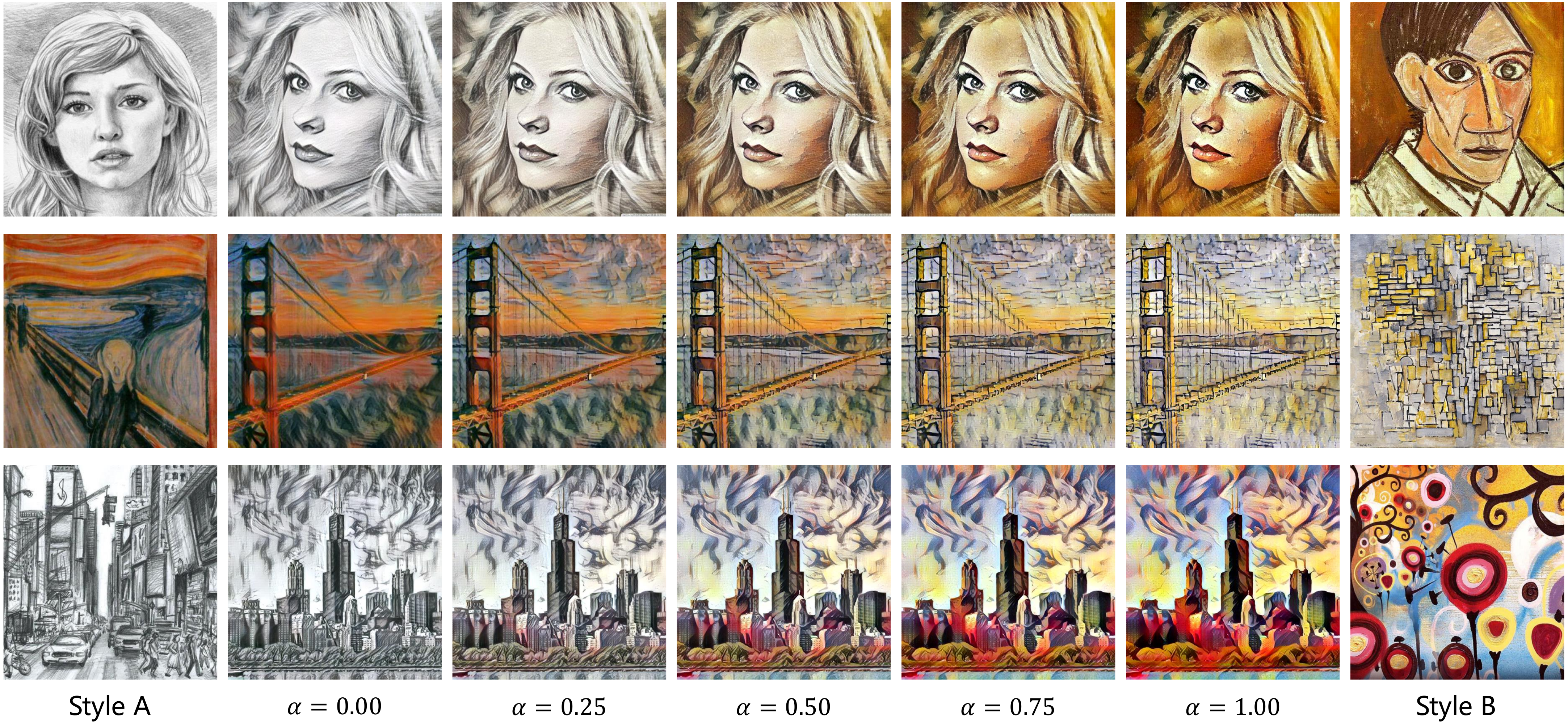}
  \caption{Style interpolation results of CAP-VSTNet on artistic style transfer.}
  \label{sp3-1-2}
\end{figure*}

We investigate the linear interpolation of extracted style representations by CAP-VSTNet. Considering the feature covariance matrices $\Sigma_{A}$ and $\Sigma_{B}$ of style images $I_{A}$ and $I_{B}$, the interpolated $\Sigma_{s}$ should be:
\begin{equation}
  \Sigma_s = (1-\alpha) \Sigma_{A} + \alpha \Sigma_{B}
\end{equation}
where $\alpha$ is the style ratio between the two. Figure~\ref{sp3-1-1} and ~\ref{sp3-1-2} present the smooth transformation from one style image to another.

\section{Comparison with ArtFlow}
\begin{figure*}
  \centering
  \includegraphics[width=0.91\linewidth]{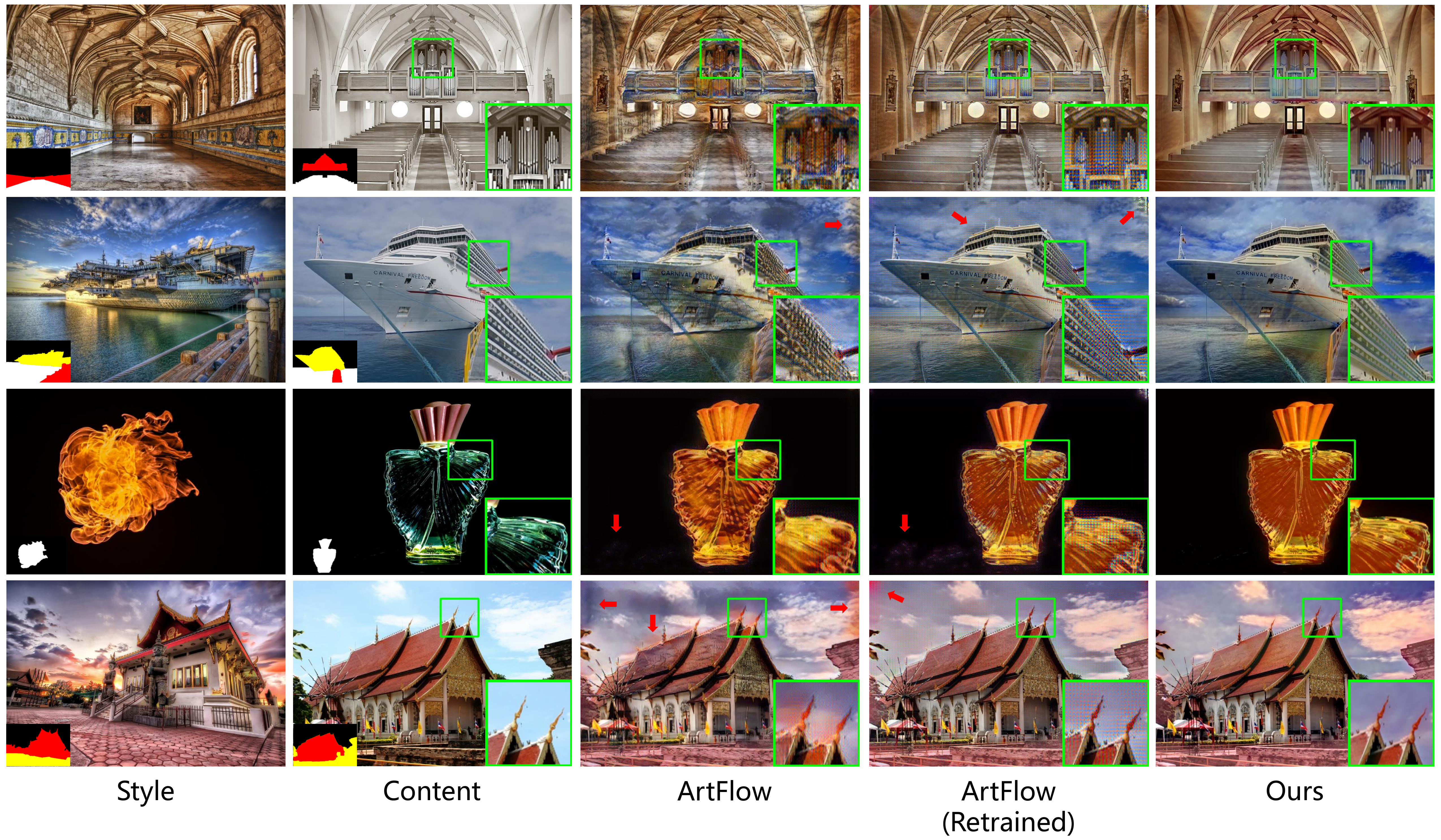}
  \caption{Visual comparisons with ArtFlow on photorealistic style transfer. We retrain ArtFlow on the Microsoft COCO (photorealistic) dataset, in order to avoid the problem of domain gap. We only increase the content loss weight to preserve more content.}
  \label{sp3-2-1}
\end{figure*}

\begin{figure*}
  \centering
  \includegraphics[width=0.91\linewidth]{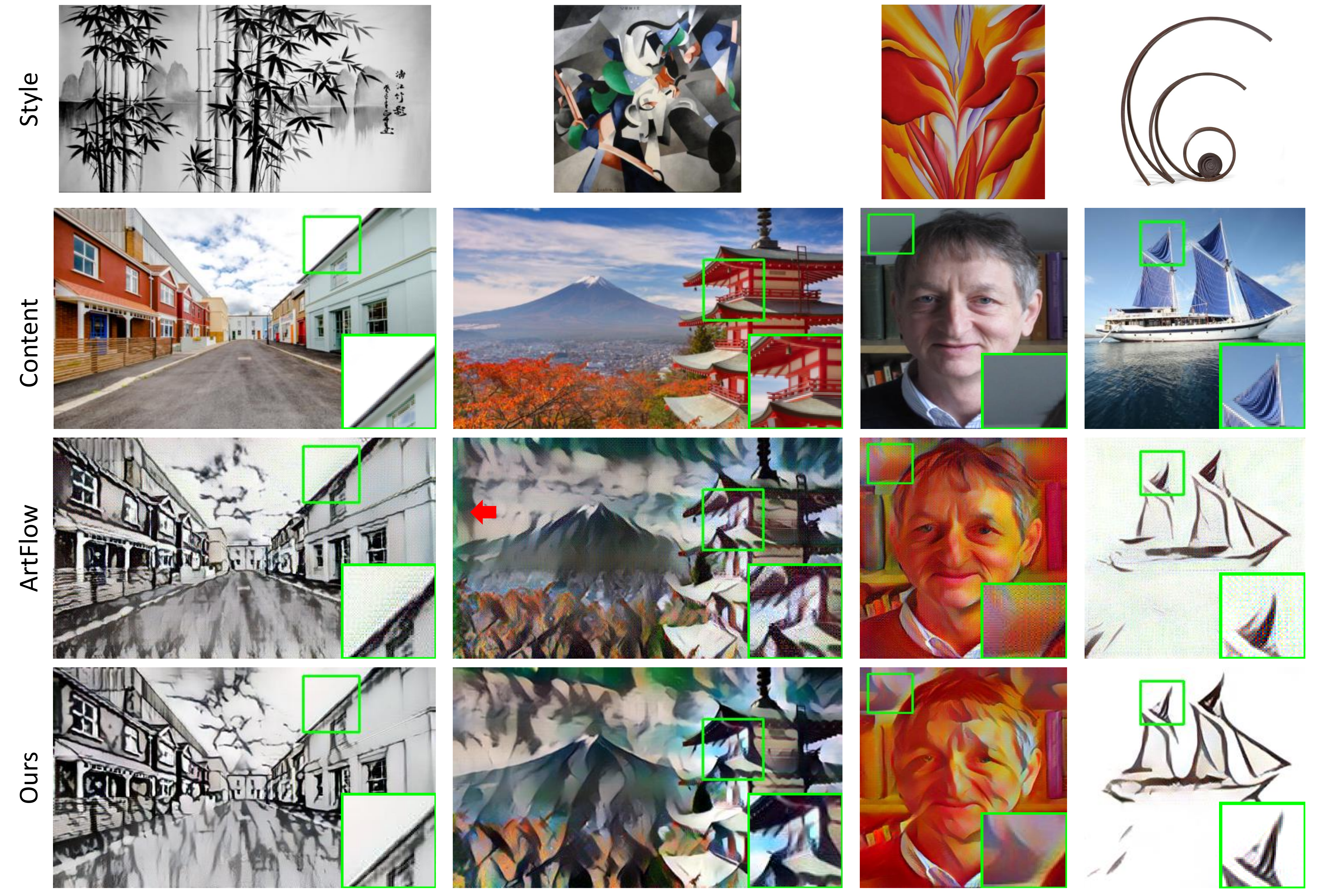}
  \caption{Visual comparisons with ArtFlow on artistic style transfer.}
  \label{sp3-2-2}
\end{figure*}

Figure~\ref{sp3-2-1} and ~\ref{sp3-2-2} show the comparisons with ArtFlow.


\section{Photorealistic Image Style Transfer}
\begin{figure*}
  \centering
  \includegraphics[width=1.0\linewidth]{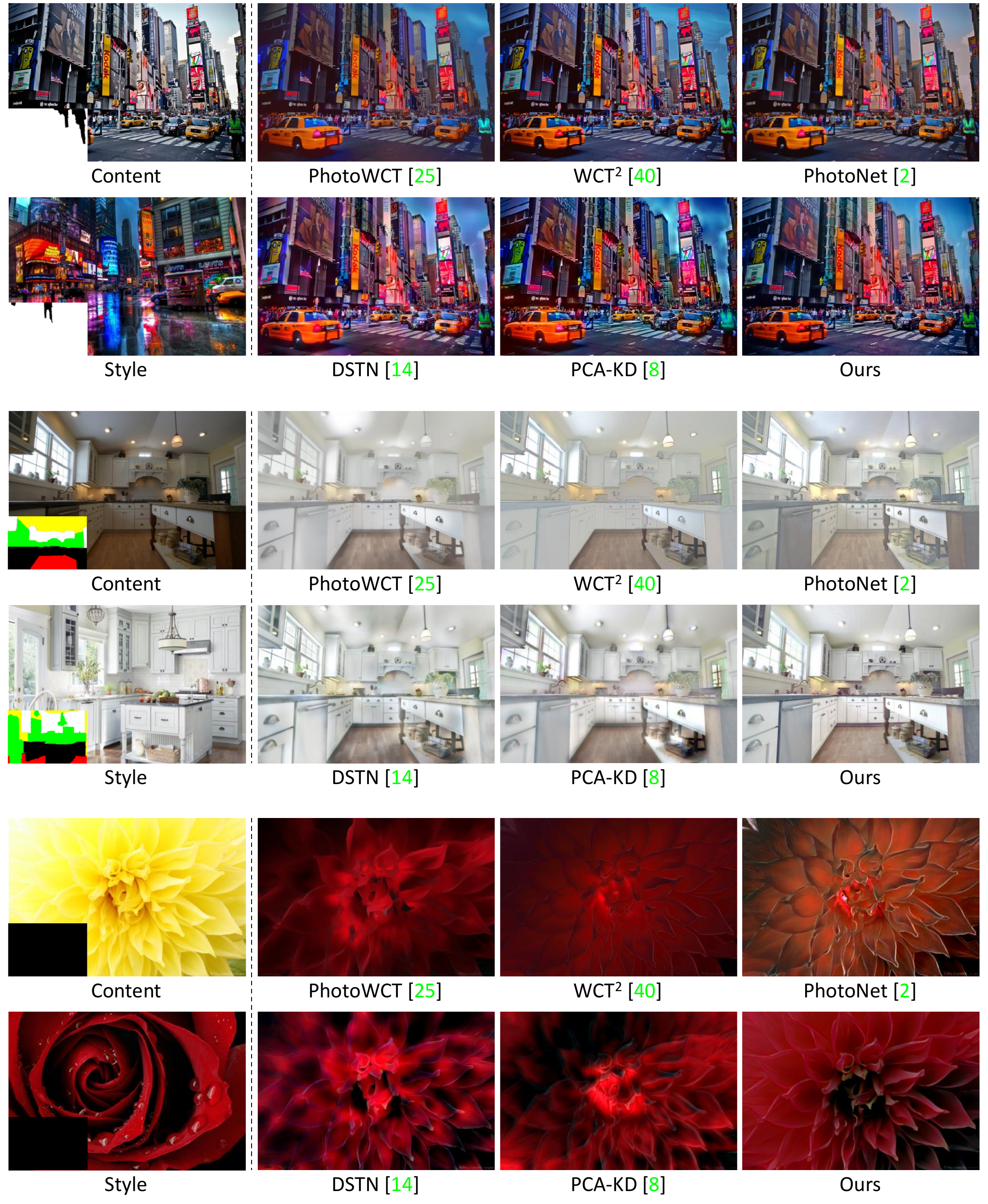}
  \caption{Visual comparisons of photorealistic image style transfer.}
  \label{sp2-1}
\end{figure*}

\begin{figure*}
  \centering
  \includegraphics[width=1.0\linewidth]{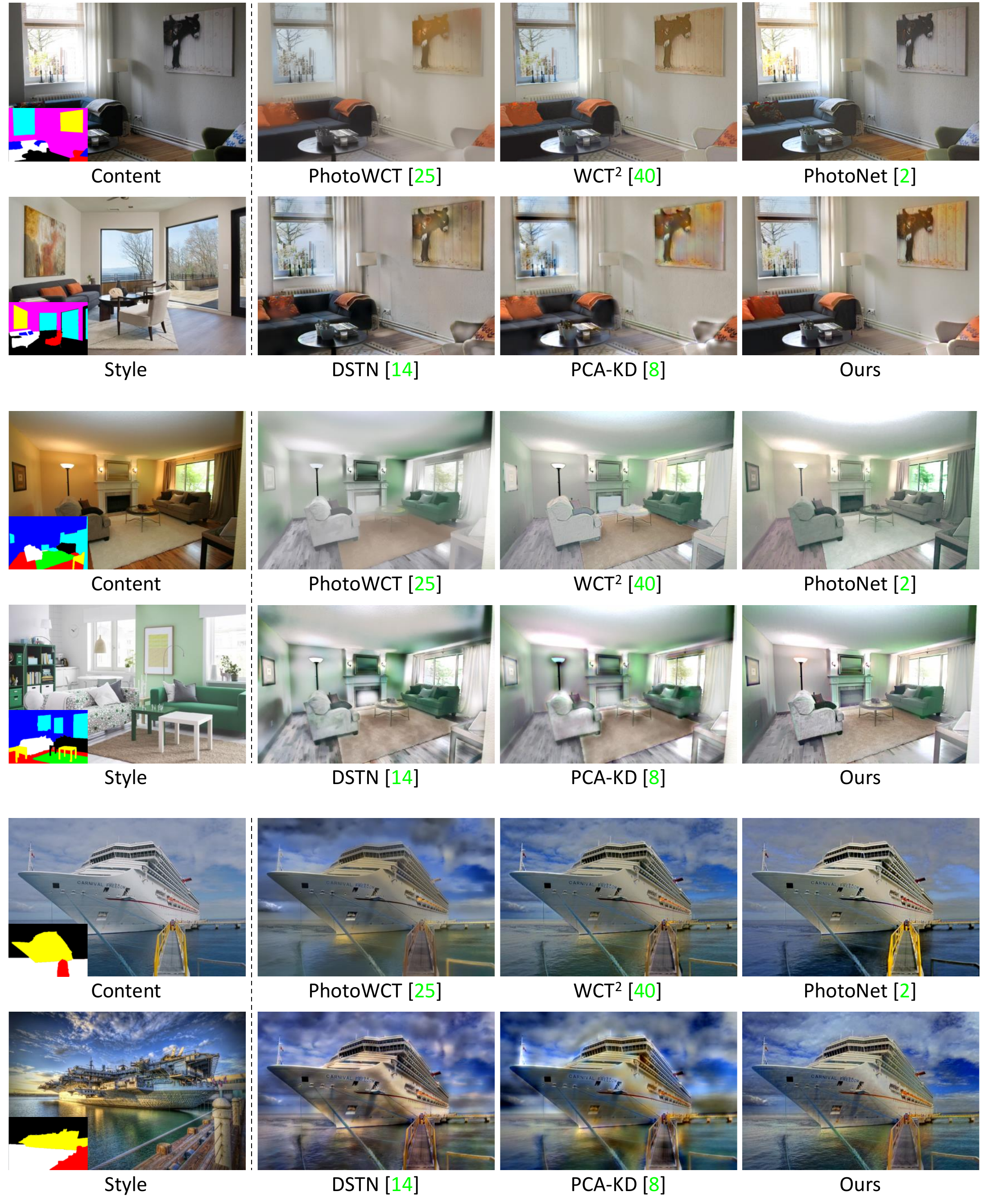}
  \caption{Visual comparisons of photorealistic image style transfer.}
  \label{sp2-2}
\end{figure*}

Figure~\ref{sp2-1} and ~\ref{sp2-2} show the comparisons with advanced photorealistic image style transfer methods.


\section{Video Style Transfer}
\begin{figure*}
  \centering
  \includegraphics[width=1.0\linewidth]{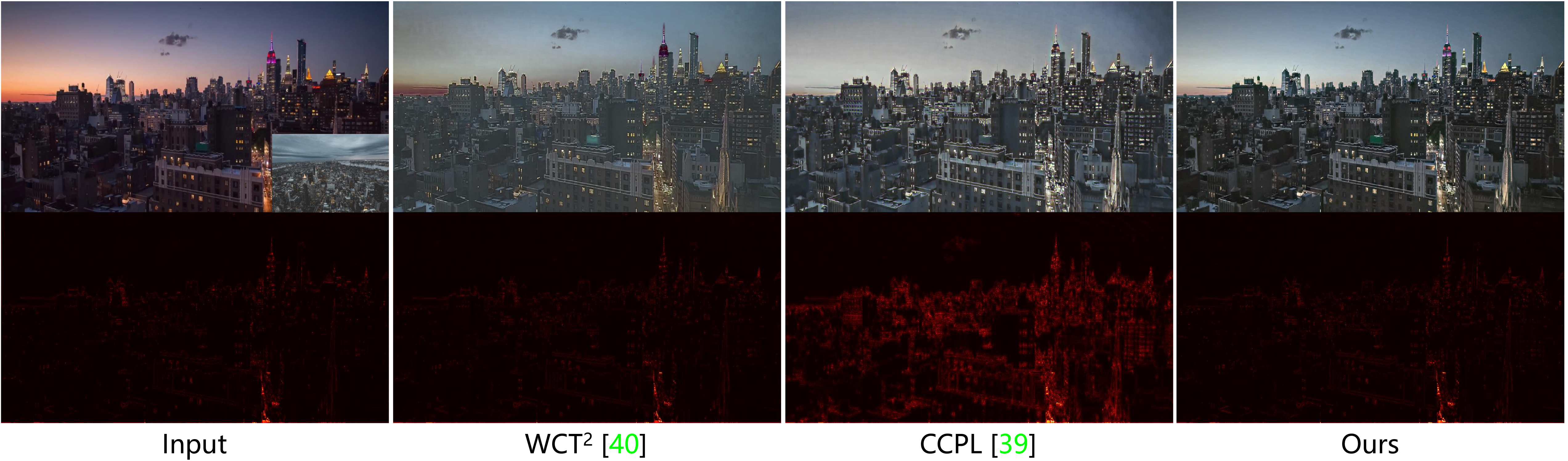}
  \caption{Visual comparison of photorealistic video style transfer. The odd rows show the stylization effect. The even rows show the temporal error heatmap of adjacent frames.}
  \label{sp4-1-1}
\end{figure*}

\begin{figure*}
  \centering
  \includegraphics[width=1.0\linewidth]{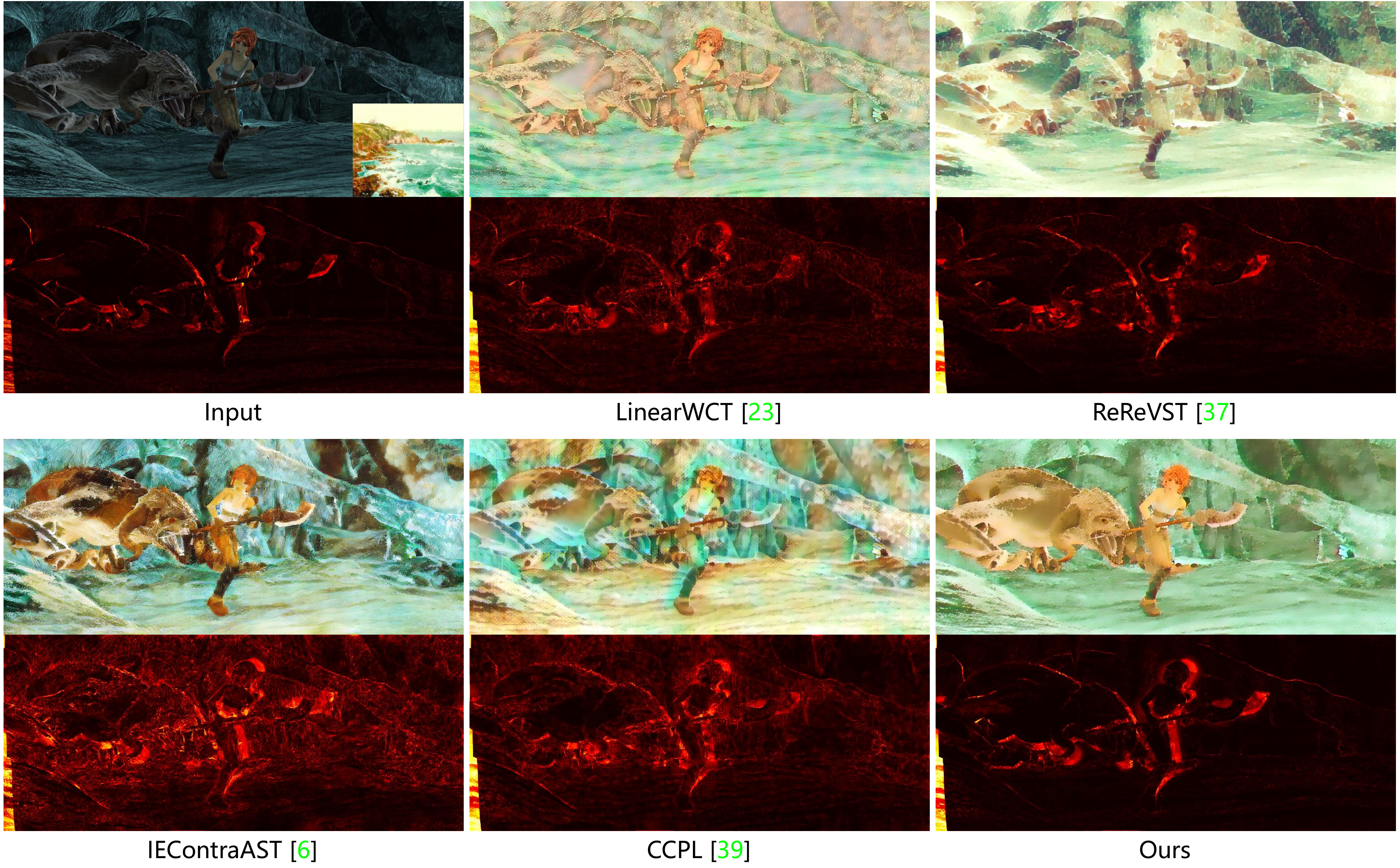}
  \caption{Visual comparison of artistic video style transfer. The odd rows show the stylization effect. The even rows show the temporal error heatmap of adjacent frames.}
  \label{sp4-2-1}
\end{figure*}

\begin{figure*}
  \centering
  \includegraphics[width=1.0\linewidth]{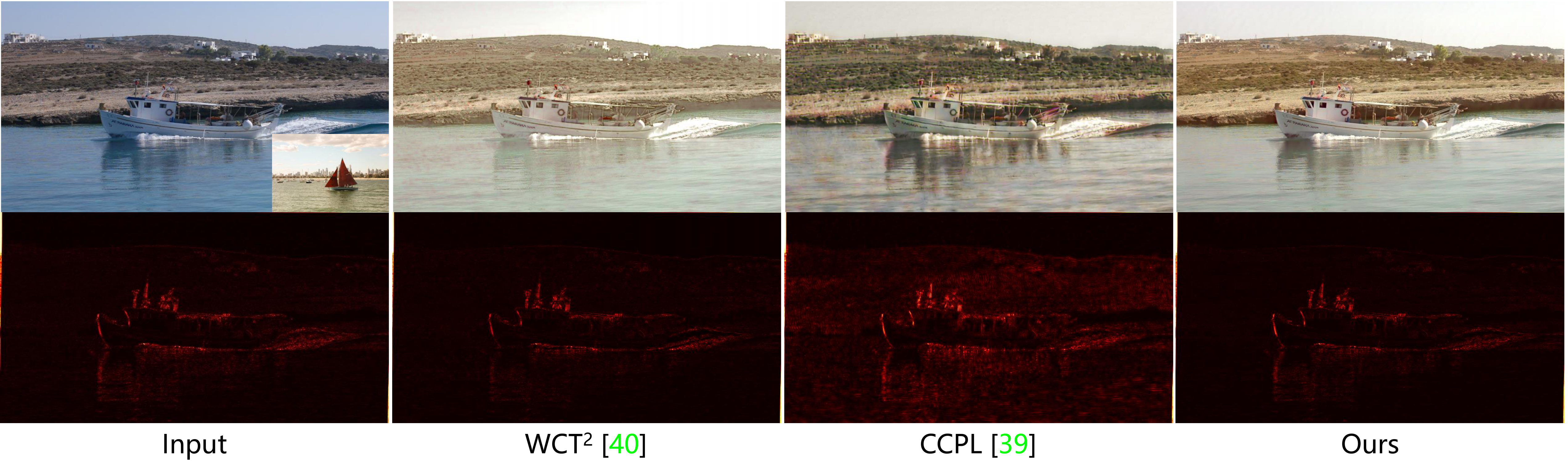}
  \caption{Visual comparison of photorealistic video style transfer. The odd rows show the stylization effect. The even rows show the temporal error heatmap of adjacent frames.}
  \label{sp4-1-2}
\end{figure*}

\begin{figure*}
  \centering
  \includegraphics[width=1.0\linewidth]{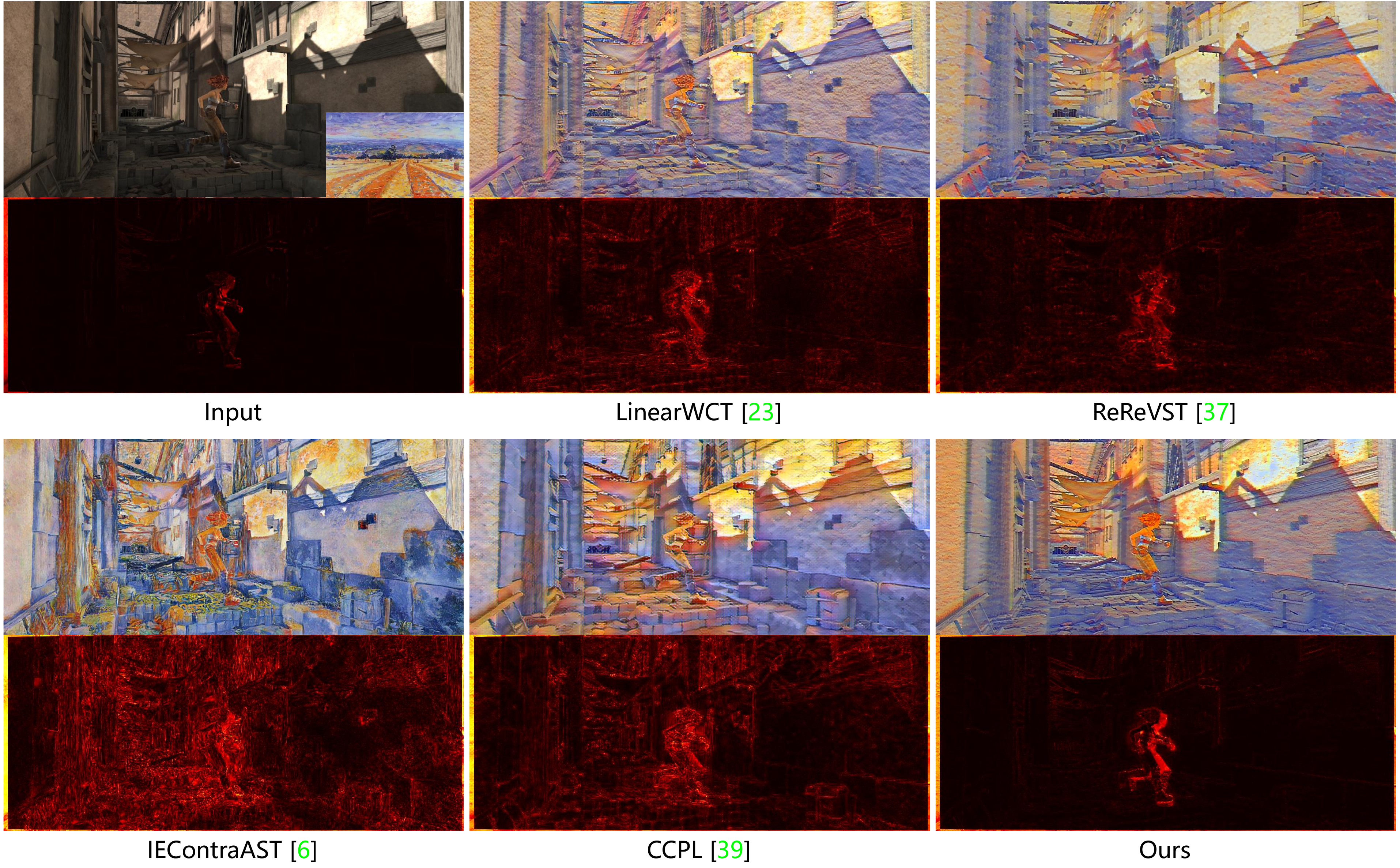}
  \caption{Visual comparison of artistic video style transfer. The odd rows show the stylization effect. The even rows show the temporal error heatmap of adjacent frames.}
  \label{sp4-2-2}
\end{figure*}

Figure~\ref{sp4-1-1}, \ref{sp4-2-1}, \ref{sp4-1-2} and \ref{sp4-2-2} show the comparisons with advanced photorealistic video and artistic video style transfer methods.

\section{Ultra-Resolution Photorealistic Style Transfer}
\begin{figure*}
  \centering
  \includegraphics[width=1.0\linewidth]{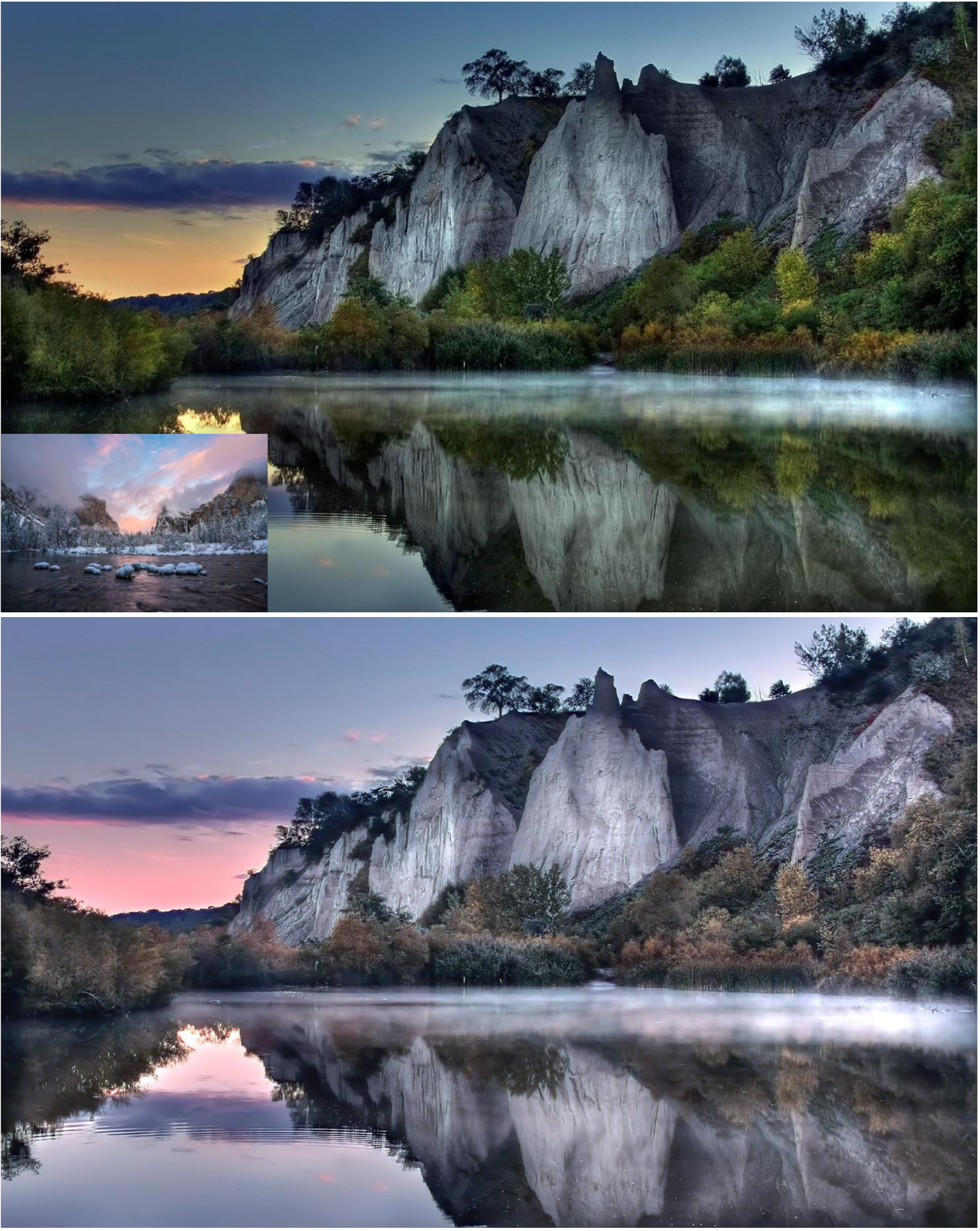}
  \caption{An example of ultra-resolution (4K) photorealistic style transfer generated by CAP-VSTNet.}
  \label{sp5-1}
\end{figure*}

\begin{figure*}
  \centering
  \includegraphics[width=1.0\linewidth]{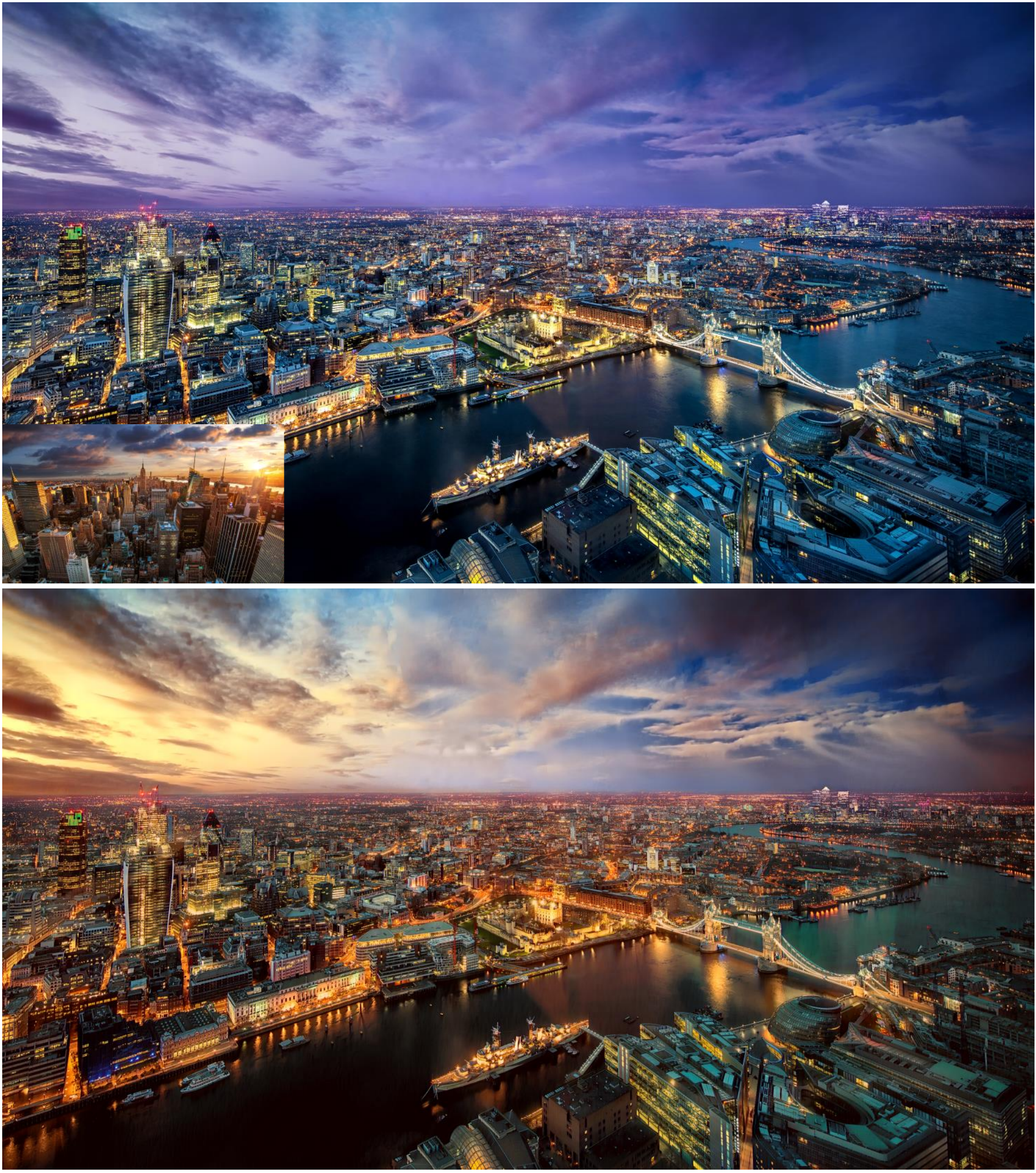}
  \caption{An example of ultra-resolution (4K) photorealistic style transfer generated by CAP-VSTNet.}
  \label{sp5-2}
\end{figure*}

Figure~\ref{sp5-1} and ~\ref{sp5-2} show the ultra-resolution (4K) photorealistic stylization results of CAP-VSTNet.



\clearpage

\clearpage

\clearpage


\end{document}